%% file: main.tex
\documentclass{article}
\usepackage{fullpage}

\usepackage{watml}
\usepackage{longtable}
\usepackage[table,x11names]{xcolor}
\usepackage{colortbl}
\usepackage{diagbox}

\renewcommand{\epsilon}{\varepsilon}

\newcommand{\XPA}{\texttt{DPS}\xspace}
\newcommand{\XPAfull}{dominant prediction score\xspace}
\newcommand{\CPA}{\texttt{EPA}\xspace}
\newcommand{\CPAfull}{ergodic prediction accuracy\xspace}
\title{Are Targeted Data Poisoning Attacks as Effective as We Think?\footnote{Equal contribution. $^\dagger$Corresponding author: \texttt{yiwei.lu@uottawa.ca}.}}

\author{%
  William Xu$^{1*}$ \quad Chenyu Zhang$^{2*}$ \quad Yihan Wang$^{2}$ \quad
  Matthew Y.~R.~Yang$^{3}$ \quad Zuoqiu Liu$^{4}$ \\[0.3em]
  Yaoliang Yu$^{2,5}$ \quad Gautam Kamath$^{2,5}$ \quad Yiwei Lu$^{5,6\dagger}$ \\[0.6em]
  $^1$Waabi AI \quad
  $^2$University of Waterloo \quad
  $^3$Carnegie Mellon University \\[0.2em]
  $^4$Google \quad
  $^5$Vector Institute \quad
  $^6$University of Ottawa
}

\begin{document}

\maketitle

\begin{abstract}
Targeted data poisoning attacks manipulate model predictions on specific test samples by injecting malicious data into training. Yet existing evaluations report average attack success rates over randomly selected targets, obscuring true worst-case effectiveness. We argue that the right evaluation focuses on the \emph{hardest} samples to poison. The same reasoning applies to defense: since targeted attacks leave no footprint at the distribution level, defenders should proactively identify the \emph{most vulnerable} samples and apply targeted countermeasures.
Given a test dataset, this paper identifies both the easiest and hardest to poison examples based on only clean model information. Specifically, we offer coarse evaluations using clean training dynamics, and fine-grained classification on poison class using poison distances and budgets. Our experiments show these metrics reliably stratify samples by poisoning vulnerability, enabling both rigorous worst-case evaluation and proactive vulnerability-aware defense.
\end{abstract}

\input{sections/introduction}

\input{sections/background}

\input{sections/method}

\input{sections/experiments}

\input{sections/ablation}

\input{sections/conclusion}

\printbibliography[title=References]

\appendix
\include{sections/appendix}

\end{document}

%% file: sections/introduction.tex
\section{Introduction}

In the past decade, machine learning (ML) models have achieved great success in various domains, largely due to the vast amount of training data available on the internet. However, this reliance on massive training datasets not only increases computational costs but also introduces significant security vulnerabilities during the data collection process \citep{SzegedyZSBEGF13,KumarNLMGCSX20}.  Adversaries can exploit these vulnerabilities through data poisoning attacks which deliberately inject malicious samples into training data either actively or passively ~\citep{GaoBBGHFPHTN20,Wakefield16, ShejwalkarHKR21,LyuYY20,carlini2024poisoning}. These attacks are particularly concerning because they can compromise model integrity at its foundation, affecting all downstream applications and users of the poisoned model~\citep{Goldblumetal21}.

Targeted data poisoning attacks  represent a specialized form of this threat, where attackers aim to manipulate model behavior for specific test instances while maintaining normal performance on all other inputs \citep[\eg,][]{ShafahiHNSSDG18,AghakhaniMWKV20,GuoL20,ZhuHLTSG19e, GeipingFHCTMG20}. We primarily focus on classification models (and show in Appendix~\ref{app:diffusion} that poisoning difficulty is similarly non-uniform in generative models), where the objective is to misclassify a particular sample to a predetermined class while maintaining correct predictions for all other inputs. Such attacks are difficult to detect as they leave little evidence in overall model performance metrics.

Current evaluations of targeted attack threats typically rely on randomly selected test samples and report average attack success rates \citep[\eg,][]{AghakhaniMWKV20,GeipingFHCTMG20} --- a methodology that conflates easy and hard targets and obscures true attack effectiveness. Security guarantees are inherently worst-case: an attack that succeeds only on easy targets does not reflect the real threat. Conversely, from the defender's perspective, since targeted attacks leave no footprint at the distribution level, the practical defense is to identify the most vulnerable samples in advance and apply targeted countermeasures. This paper addresses both angles by asking: \emph{can we reliably identify the hardest and easiest samples to poison, using only clean model information?}

We answer this question by introducing two levels of metrics. At the coarse level, \CPAfull (\CPA) and its label-agnostic surrogate \XPAfull (\XPA) leverage clean training dynamics to broadly distinguish easy-to-poison from hard-to-poison samples. At the fine-grained level, poisoning distance $\delta$ and poison budget lower bound $\tau$ provide poison-class-specific predictions, capturing which poison class is easiest to induce for a given sample. Importantly, all metrics are computable without executing any actual attacks: the coarse metrics rely solely on clean training dynamics, while the fine-grained metrics require only the clean model weights.

Our experimental results confirm the effectiveness of both levels of metrics. At the coarse level, \CPA reliably separates easy- and hard-to-poison samples: for samples the model is highly confident about during clean training, the average attack success rate drops to as low as 47.28\% for certain target classes (vs.\ 93.83\% for low-\CPA samples), revealing that existing attacks are substantially less effective than average-case evaluations suggest. \XPA provides a label-agnostic surrogate with consistent behavior. At the fine-grained level, poisoning distance $\delta$ and poison budget lower bound $\tau$ further resolve which poison class is easiest to induce for a given sample, providing poison-class-specific predictions that \CPA alone cannot capture.

In summary, our work makes three distinct contributions: (1) We argue that existing evaluations of targeted data poisoning attacks are misleading, as average attack success rates over randomly selected samples conflate easy and hard targets; a more informative evaluation instead focuses on the hardest samples to poison. (2) We introduce two levels of metrics computable without executing any attacks: coarse metrics (\CPAfull and \XPAfull) derived from clean training dynamics, and fine-grained, poison-class-specific metrics (poisoning distance $\delta$ and budget lower bound $\tau$) requiring only clean model weights. (3) Our experiments demonstrate that these metrics reliably identify the easiest- and hardest-to-poison samples, enabling both rigorous worst-case evaluation and proactive vulnerability-aware defense.

%% file: sections/background.tex
\section{Background}

\paragraph{Threat model and notations:}

We first specify our threat model and list our notations below.
\begin{itemize}[leftmargin=*]
    \item \textbf{Setup}: An adversary replaces\footnote{We further discuss adding attacks in Appendix \ref{app:add_replacing}.} part of the clean training set $\Dcl$ to a poisoned set $\Dpo$, such that the defender trains on the modified training set $\Dtr$ and deploys the model on a test set $\Dte$ containing $\xte$. The goal is to \emph{alter the prediction} of a specific test sample $\xte$\footnote{The attacker cannot modify $\xte$ directly, but indirectly changes the model's behavior by deploying $\Dpo$ during training. This is a key difference from adversarial examples, which directly modify $\xte$ without any poisoned set or retraining.} from its true class $y_t$ to a specific poison class $y_p$, while leaving all other predictions intact.
    \item \textbf{Attacker}: We consider a \emph{white-box attack},\footnote{Attacks could also be performed in a partially \emph{black-box} fashion using surrogate models, but such attacks suffer a severe performance drop \citep{SchwarzschildGGDG20}. To measure the strongest possible threat from the defender's perspective, we consider the white-box setting.} where the attacker is aware of $\Dcl$, the model architecture, the training scheme, and the inclusion of $\xte$ in the test set. The poisoning budget is $\epsilon = \frac{|\Dpo|}{|\Dtr|}$, typically low (e.g., $\epsilon=1\%$). We further constrain $\Dpo$ to contain only clean-labeled poison data, i.e., elements of $\Dpo$ are generated by adding human-imperceptible perturbations (e.g., with $\ell_\infty$ constraints) to clean training images without changing their original labels.
    \item \textbf{Evaluation}: An attack is successful when $f(\xte; \wv_p) = y_p$. Note that this is strictly stronger than misclassification, as it requires prediction of a specific target label. The defender has access to $\Dcl$ and can train a clean model $\wv_c$ on $\Dcl$ alone, which serves as the basis for computing our difficulty metrics without requiring any knowledge of the attack.
\end{itemize}

\textbf{Other notation:} We denote the clean model parameters (a model $f$ trained only on $\Dcl$) to be $\wv_c$ and poisoned model parameters to be $\wv_p$. Let $\ell(\zv;\wv)$ be our loss that measures the fitness of model $\wv$ on data $\zv \in \mathcal{Z}$, \eg, $\zv = (\xv, y)$. Let $\gv(\zv) = \gv(\zv; \wv) = \nabla_{\wv} \ell(\zv; \wv)$ be the gradient vector with respect to a fixed model $\wv$ evaluated at data $\zv$.

\textbf{Targeted data poisoning:}
In this paper, we focus on targeted data poisoning attacks \citep[\eg,][]{ShafahiHNSSDG18,AghakhaniMWKV20,GuoL20,ZhuHLTSG19e} that affect only specific test samples and discuss other types of poisoning attacks in Appendix \ref{app:data-poisoning}.
Given a test sample $(\xte, y_t)$, the problem can be formulated into the bi-level optimization problem below:
\begin{align}
    \min_{\Dpo}~ \ell((\xte,y_p), \wv_*),  ~\mathrm{s.t.}~\wv_* \in \argmin_{\wv} ~\ell(\Dtr, \wv),
\end{align}
where the attacker aims to enforce the prediction of $\xte$ to be $y_p$ through crafting and injecting $\Dpo$ into the training set $\Dtr$. This problem is hard to solve directly as the outer minimization problem depends on $\Dpo$ only implicitly through the solution $\wv_*$ of the inner problem. Existing attacks consider relaxations of this primal problem, for example, a fixed feature extractor \citep{AghakhaniMWKV20,ZhuHLTSG19e} or approximating the gradient of target parameters \citep{ShafahiHNSSDG18}. As our poisoning difficulty metrics do not depend on the specific design of attack algorithms, we omit the attack details and refer readers to the above references.
We \emph{do not} consider backdoor attacks \citep{GuDG17,ChenLLLS17,SahaSP20,TranLM18}, as their difficulty largely depends on the choice of trigger rather than properties of the target sample itself.

%% file: sections/method.tex
\section{Metrics for Targeted Data Poisoning Difficulty}

\begin{figure}
  \begin{center}
    \vspace{-1em}\includegraphics[width=0.5\linewidth]{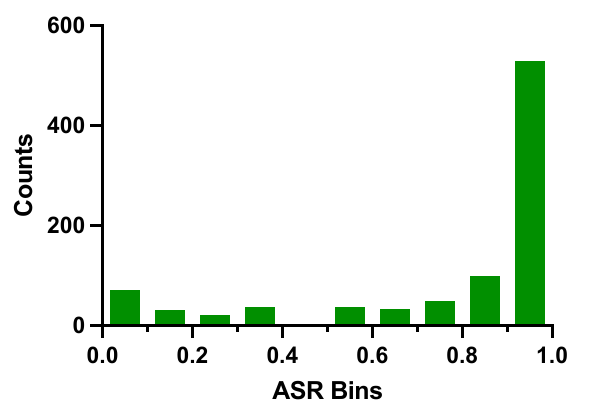}
  \end{center}
  \vspace{-1em}
  \caption{Histogram of ASR of gradient matching on 100 test samples in the class ``plane'' in CIFAR-10. While targeted attacks are generally effective, there is substantial variance across samples.}
\label{fig:motivation_target_attacks}
\vspace{-1em}
\end{figure}

To illustrate the variance of attack performance across test samples, we apply gradient matching \citep{GeipingFHCTMG20} to the first 100 test samples in the class ``plane'' of CIFAR-10 \citep{Krishevsky09}, targeting all nine other poison classes (900 attacks in total). For each attack, we perform 8 independent trials with $\epsilon=1\%$ and report the attack success rate as a histogram in \Cref{fig:motivation_target_attacks}.
We observe that while the average attack ASR is 78.94\%, this masks the fact that a slight majority (58.67\%) is very easy to poison (ASR = 100\%), and that 17.44\% is difficult to poison (ASR < 40\%).
This variance has practical implications for both sides: from the attacker's perspective, average ASR over random samples obscures which samples are truly hard to compromise; from the defender's perspective, not all samples require equal scrutiny, and identifying the most vulnerable ones enables prioritized, targeted protection. We will show in \Cref{sec:exp} that the choice of poison class also introduces significant variance.

We introduce two levels of metrics for quantifying targeted data poisoning difficulty. At the coarse level, \CPA and \XPA (\Cref{sec:cpa}) leverage clean training dynamics to broadly identify easy- and hard-to-poison samples, enabling defenders to prioritize protection and evaluators to select informative worst-case targets. At the fine-grained level, poisoning distance $\delta$ and poison budget lower bound $\tau$ (\Cref{sec:delta}) provide poison-class-specific predictions for a given target sample $\xte$. A key strength of both levels is that they are computable without executing any actual attacks.

\subsection{Coarse Metrics: Clean Training Dynamics}
\label{sec:cpa}
A natural indicator of a test sample's robustness to poisoning is how consistently the model classifies it throughout training. Intuitively, a sample that is classified correctly and confidently across many random initializations and training epochs is unlikely to be destabilized by a small number of poison samples. Conversely, a sample whose predicted label fluctuates during training already sits near the boundary of the model's decision regions, making it inherently easier to push toward a different class through poisoning. This motivates the following hypothesis:

\begin{hypothesis}
  The classification difficulty of a test sample $\xte$ is negatively correlated with its poisoning difficulty, i.e., a sample $\xte$ that is easy to classify is correspondingly difficult to poison. 
  \label{assump:classification}
\end{hypothesis}

To verify the above hypothesis, it is necessary to establish a robust measure of classification difficulty, which we approach by examining prediction behavior throughout training.
\begin{definition}[Ergodic Prediction Accuracy, \CPA] 
We say the classification difficulty for a target sample $\xte$ can be measured by the ergodic average correctness (denoted by the indicator function) for $N$ training epochs with $M$ different initializations:
\begin{align}
   \emph{\CPA} = \cfrac{1}{MN} \sum_{m=1}^M\sum_{n=1}^{N} \mathds{1}\{f_{m,n}(\xte) = y_t\},
   \label{eq:cpa}
\end{align}
\label{def:cpa}
\end{definition}

When the model update is ergodic \citep{RoydenFitzpatrick10} with large $M$ and $N$, \CPA converges to $\Pr[ f(\xte; \wv^*) = y_t] $ where $\wv^*$ follows the invariant distribution of the update process. 

Although we use the hard prediction of $f_{m,n}(\xte)$ above, \CPA can also be calculated using the model's softmax outputs as a soft measure of confidence.\footnote{\CPA is only one possible way to measure classification difficulty, we discuss alternative approaches in Appendix \ref{app:class_diff}.} We demonstrate the performance of both approaches in \Cref{sec:exp}. We emphasize a key limitation of \CPA: it requires access to the true test label $y_t$, which is not always practical. We address this with a label-agnostic surrogate:

\begin{definition}[Dominant Prediction Score, \XPA] We say classification difficulty can be measured by the label \(y \in \mathcal{Y}\) which dominates the prediction, where \(\mathcal{Y}\) is the set of all class labels:
\begin{align}
   \emph{\XPA} = \max_{y \in \mathcal{Y}} \cfrac{1}{MN} \sum_{m=1}^M\sum_{n=1}^{N} \mathds{1}\{f_{m,n}(\xte) = y\},
   \label{eq:xpa}
\end{align}
\label{def:xpa}
\end{definition}
A higher \XPA indicates that predictions are concentrated on a single label, suggesting that $\xte$ is easier to classify and harder to poison, and vice versa. We empirically demonstrate that \XPA is a good surrogate for \CPA without access to $y_t$ in \Cref{sec:exp}.

\subsection{Fine-grained Metrics: Poisoning Distance and Budget}
\label{sec:delta}

\begin{figure}{h}
    \centering
    \includegraphics[width=0.5\linewidth]{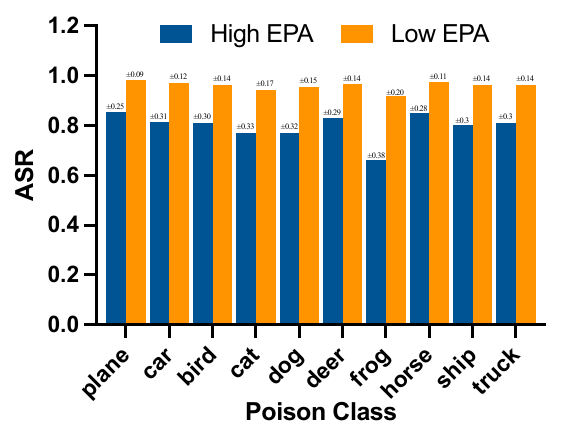}
    \vspace{-.5em}
    \caption{Attack success rates (ASR) of gradient matching on CIFAR-10 across different poison classes $y_p$ for the 50 highest and 50 lowest \CPA samples. ASR varies significantly across different $y_p$, motivating the need for fine-grained metrics.}
    \label{fig:main_poison_class_differences_high_and_low_cpa}
\end{figure}

While \CPA and \XPA provide coarse indicators of poisoning difficulty, they are agnostic to the choice of poison class $y_p$. As shown in \Cref{fig:main_poison_class_differences_high_and_low_cpa}, the attack success rate (with $\epsilon=1\%$) varies significantly across different poison classes for the same target sample, motivating the need for poison-class-specific metrics.

\paragraph{1. Poisoning distance $\delta$} We consider the ultimate goal of targeted poisoning:

\emph{\textbf{Goal 1}: An adversary aims at modifying model parameters from clean parameters $\wv_c$ to poisoned ones $\wv_p$ such that $f(\xte;\wv_p)=y_p$}. 

Data poisoning implements an indirect way to achieve this goal through crafting $\Dpo$ and training on $\Dcl \cup \Dpo$. Goal 1 enables us to measure poisoning difficulty by comparing $\wv_p$ and $\wv_c$ directly. Specifically, we propose a hypothesis on poisoning distance:
\begin{hypothesis}
For a target sample $\xte$ and poison class $y_p$, the distance $\delta = d(\wv_c, \wv_p)$ between a clean model $\wv_c$ and the closest model $\wv_p$ satisfying $f(\xte;\wv_p)=y_p$ is positively correlated with poisoning difficulty, i.e., a larger $\delta$ indicates that $\xte$ is more difficult to poison into class $y_p$.
\label{assump:delta}
\end{hypothesis}
Note that 
$d(\cdot)$ is a distance function that we will specify soon. Here $\delta$ is a sample-wise metric as its calculation depends on the sample-specific poisoned parameter $\wv_p$. Moreover, \Cref{assump:delta} naturally considers the choice of $y_p$, which is embedded in the definition of $\wv_p$.

However, from a defender's perspective, validating \Cref{assump:delta} directly is non-trivial as the calculation of $\delta$ depends on the poisoned parameters $\wv_p$, which are unknown without performing an actual attack. Luckily, data poisoning is not the only viable way to achieve Goal 1, and we propose a proxy to generate $\wv_p$ and measure $\delta$ without performing any data poisoning attacks:
\begin{definition}[Poisoning Distance $\delta$] Starting from a clean model $\wv_c$, we say the poisoning distance is the smallest step size required to modify $\wv_c$ in one step such that the model classifies $\xte$ as $y_p$:
\begin{align}
    \delta = \min\{\eta>0: f(\xte; \wv_c-\eta\cdot\gv)=y_p\},
\end{align}
where $\gv=\nabla_{\wv} \ell(\xte; \wv_c, y_p)$ is the gradient of the loss with respect to model parameters, evaluated at $\wv_c$ with target label $y_p$. We also obtain our proxy $\wv_p = \wv_c-\eta\cdot\gv$.
\label{def:distance}
\end{definition}

While such a proxy may be different from a real data poisoning attack, $\delta$ intuitively measures the efforts needed to achieve the attack goal from a gradient perspective.\footnote{We note that our core idea of poisoning distance is closely related to model-targeted indiscriminate attacks and we extend our discussion in Appendix \ref{app:MTA}.}
We provide a simple binary search estimator for $\delta$ in \Cref{alg:delta} in Appendix \ref{app:algorithms}. One advantage of $\delta$ over \CPA is that it does not depend on the training process or the clean data $\Dcl$, making it practical for users who have access only to pre-trained model weights, for example when assessing the vulnerability of their own data against foundation models.

\paragraph{2. Poison budget lower bound $\tau$} Aside from poisoning distance, a complementary fine-grained metric measures poisoning difficulty through the poison budget $\epsilon$. It is clear that an attack is easier if less poisoned data is required, i.e., a lower $\epsilon$ suggests an easier attack. Here we aim to answer an intriguing question:
\emph{Is it possible to measure the lowest $\epsilon$ needed to poison a model such that a given test sample $\xte$ is misclassified as $y_p$, without performing any attacks?}

Conveniently,  \citet{LuKY23} provides theoretical tools for measuring the (relative) number of poisoned samples $|\Dpo|$ needed to reach some target parameters $\wv_p$ (e.g., the proxy we generated in \Cref{def:distance}), i.e., the role of poison budget $\epsilon$. Specifically, \cite{LuKY23} provides a lower bound with respect to $\epsilon$ on poisoning reachability. 
We present a simplified version of their results:

\begin{theorem}[Poisoning reachability, Theorem 2 of~\citet{LuKY23}] Given a classification task with $c$ classes and a set of target parameters $\wv_p$, $\wv_p$ is poisoning reachable (defined by vanishing gradient over training on $\Dcl \cup \Dpo$) only if the condition below holds (necessary condition):\footnote{Note that Theorem 2 in \citet{LuKY23} presents poisoning reachability for binary linear models, and we consider the general form in Equation (10) on multiclass neural networks.}
\begin{align}
    \epsilon \geq \tau: = \max\left\{\cfrac{\inner{\wv_p}{ \gv(\Dcl)}}{\Wfk(c-1/e)},0\right\},
\end{align}
where $\Wfk(\cdot)$ is Lambert's W function, $\gv(\Dcl) = \gv(\Dcl; \wv_p) = \tfrac{1}{|\Dcl|} \sum_{\zv \in \Dcl} \nabla_{\wv_p} \ell(\zv; \wv_p)$.
\label{theorem:tau}
\end{theorem}

\Cref{theorem:tau} enables us to calculate $\tau$, the lower bound of poisoning budget $\epsilon$ for a given target test sample $\xte$, the corresponding poison class $y_p$, the target parameter $\wv_p$, and the clean training set $\Dcl$. 
In \Cref{sec:exp}, we will show that $\tau$ is a direct indicator of poisoning difficulty.

%% file: sections/experiments.tex
\section{Experiments}
\label{sec:exp}

In this section, we validate our two-level framework for quantifying targeted poisoning difficulty. We first evaluate our coarse metrics (\CPA and \XPA) on identifying the hardest and easiest samples to poison, demonstrating that existing attacks are substantially less effective on high-\CPA samples than average-case evaluations suggest. We then evaluate our fine-grained metrics ($\delta$ and $\tau$) on predicting poison-class-specific difficulty. Finally, we present ablation studies on datasets, model architectures, and poisoning budgets.

\subsection{Experimental settings} 

We evaluate on the targeted attacks provided in the unified benchmark of \cite{SchwarzschildGGDG20}.\footnote{\url{https://github.com/aks2203/poisoning-benchmark}}

\textbf{Datasets \& models:} We consider classification tasks on CIFAR-10 \citep{Krishevsky09} with 10 classes, 50,000 clean training samples and 10,000 test samples in our main experiments,  and TinyImageNet \citep{le2015tiny} with 200 classes, 100,000 clean training samples, 10,000 validation samples, and 10,000 test samples in our ablation study. We apply ResNet-18 \citep{HeZRS16} for CIFAR-10 and VGG-16 \citep{simonyan2014very} for TinyImageNet.

\textbf{Training schemes:}
We consider two training schemes: (1) Training from scratch, where we initialize the model with random weights. For clean training, we use the clean training set $\Dcl$, for data poisoning we use $\Dcl \cup \Dpo$;\footnote{Note that for replacing attacks, $\Dcl$ can be changed  after poisoning, see \Cref{app:add_replacing} for discussion.} (2) Transfer learning for CIFAR-10, where we utilize a frozen model pretrained on CIFAR-100, and fine-tune an additional linear head on a subset of CIFAR-10 that contains the first 250 images per class. For both scenarios, we train the model for 40 epochs.

\textbf{Targeted attacks:} We examine three attack methods listed in the unified benchmark: (1) Gradient matching (GM) \citep{GeipingFHCTMG20}\footnote{\url{https://github.com/JonasGeiping/poisoning-gradient-matching}} for training from scratch;
(2) Feature collision (FC) \citep{ShafahiHNSSDG18} for transfer learning; and (3) Bullseye polytope (BP) \citep{AghakhaniMWKV20} for both.\footnote{We neglect Convex Polytope (CP) \citep{ZhuHLTSG19e} as it is extremely expensive. It takes 100 seconds and 40 seconds to run one attack instance for BP and FC, respectively, while it takes more than 1 hour to run CP on a NVIDIA 4090 GPU. As our experiments require thousands of attack instances, it is infeasible to run CP.} For training from scratch, we perform 8 random model initializations and calculate the attack success rate (ASR) by dividing the number of successful attacks by 8. For transfer learning, as the model initialization is mostly fixed, we only consider one attack trial each. For all attacks, unless specified otherwise, we use a poisoning budget $\epsilon=1\%$.

\textbf{Measuring poisoning difficulty:} (1) To calculate \CPA for each test sample, we train the model with $M = 100$ (CIFAR-10), and $M = 8$ (TinyImageNet) random initializations for $N=40$ epochs. We consider the model prediction for training from scratch and model confidence for transfer learning; (2) To obtain $\delta$, for each choice of $(\xte,y_p)$, we consider 8 model initializations to generate 8 $\wv_c$, apply \Cref{alg:delta} on each $\wv_c$ and obtain the average; (3) For the calculation of $\tau$, we only apply one set of $\wv_c$ and consider the number of classes $c=10$ for CIFAR-10 and $c=200$ for TinyImageNet and apply \Cref{theorem:tau}. We further report our resource and computational time in Appendix \ref{app:compute}.

\begin{figure}[h]
\vspace{-.3em}
    \centering
    \begin{subfigure}{0.41\textwidth}
        \includegraphics[width=\linewidth]{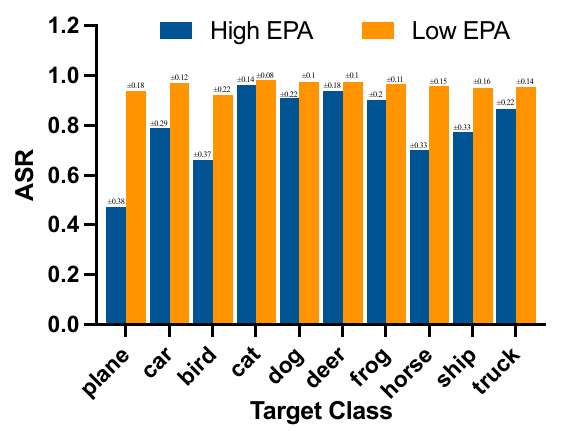}
        \vspace{-1.5em}        
        \caption{ASR for different $y_t$}
    \end{subfigure}
    \hfill
    \begin{subfigure}{0.47\textwidth}
        \includegraphics[width=\linewidth]{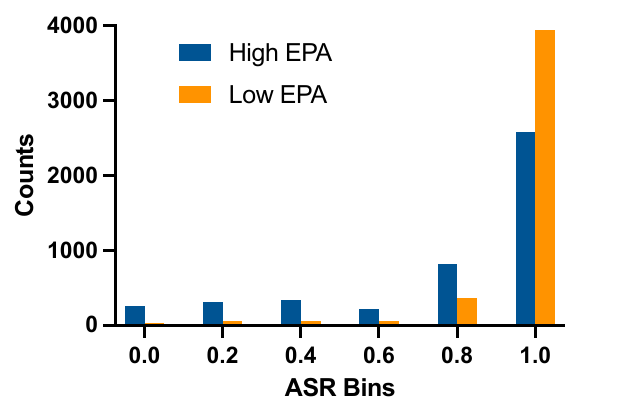}
        \vspace{-1.5em}
        \caption{ASR distribution}
    \end{subfigure}
    \vspace{-.3em}
    \caption{Poisoning difficulty stratified by \CPA on CIFAR-10 (GM, training from scratch). High-\CPA samples are substantially harder to poison than low-\CPA samples across all target classes (a), and the overall ASR distribution shifts markedly between the two groups (b).}
    \label{fig:main_cpa}
\end{figure}

\subsection{Coarse Metrics: \CPA and \XPA}
A key claim of this paper is that average-case evaluations over random target samples obscure true worst-case attack effectiveness. We test this by using \CPA to stratify test samples and comparing attack success rates between the hardest and easiest targets.

\begin{figure}
\vspace{-1em}
    \centering
    \includegraphics[width=0.85\linewidth]{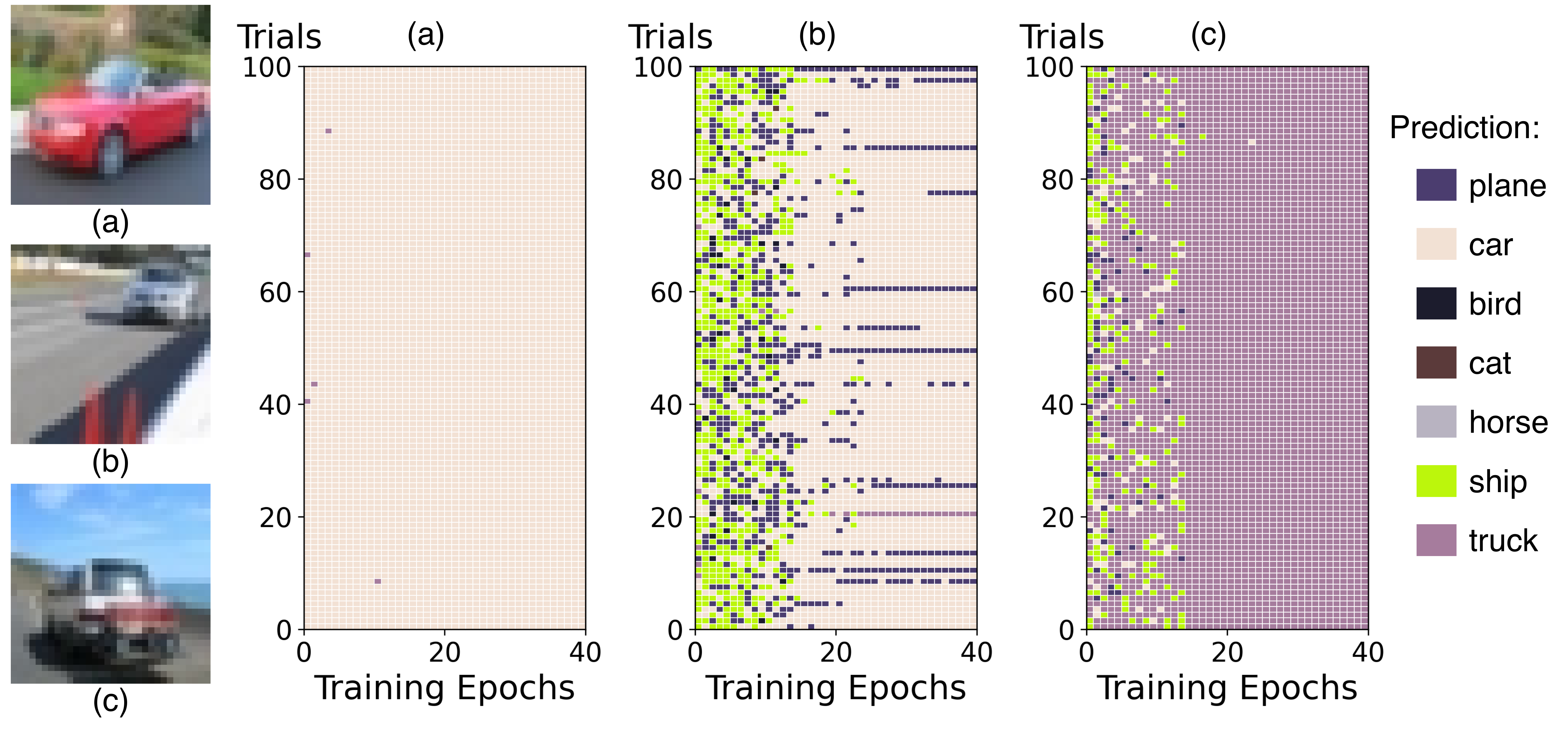}
    \caption{\CPA for three test instances in the class ``car''. Image (a): high \CPA: 0.9988; ASR: 22.22\%. Image (b): medium \CPA: 0.6775; ASR: 90.28\%. Image (c): low \CPA: 0.0275; ASR: 98.61\%.}
    \label{fig:three_heat_maps}
\end{figure}

\textbf{Training from scratch:}
For the from-scratch setting on ResNet-18/CIFAR-10, we perform clean training on $\Dcl$ with the prespecified $M$ and $N$ to identify the 50 target samples with the highest and lowest \CPA in each target class $y_t$. For each target sample, we perform the GM attack on all possible (9) poison classes $y_p$ for 8 randomly initialized model weights. We thus run $(50+50)\times 10 \times 9 \times 8 = 72000$ attack instances in total. 
In \Cref{fig:main_cpa}, we observe that \CPA is a reliable indicator of poisoning difficulty, where higher \CPA consistently yields lower ASR. The gap is substantial: for target class ``plane'', the ASR drops to $47.28\%$ for high-\CPA samples versus $93.83\%$ for low-\CPA samples. This shows that average-case evaluations — which mix hard and easy targets — significantly overstate true worst-case attack effectiveness. We note that some classes are generally more vulnerable, consistent with their lower clean test accuracy (e.g., 82.66\% for cat and 85.73\% for dog). We present a visualization of \CPA for instances with high, medium, and low \CPA values in \Cref{fig:three_heat_maps}, illustrating the range of poisoning vulnerability across samples.

\textbf{Transfer learning:}
For the transfer learning setting on ResNet-18 and CIFAR-10, we again identify the 50 target samples with the highest and lowest \CPA and apply an additional restriction that all identified target images are classified correctly at the final epoch in all $M$ clean training runs. \Cref{tab:easy_vs_hard_transfer_learning} shows our main result on CIFAR-10 for two attacks FC and BP. We observe that the average ASR for test samples with high \CPA is much lower than the ones with low \CPA for both attacks. We provide additional results on transfer learning in \Cref{add_exp:transfer}.

\textbf{\XPA as a surrogate for \CPA}:
Recall that \XPA measures the fraction of training-time predictions assigned to the \emph{dominant label} rather than the ground-truth label \(y_t\). To evaluate whether \XPA serves as a practical surrogate for \CPA, we compute \XPA for all CIFAR-10 test samples and rank them by their \XPA scores. Among the top 20\% and top 40\% of samples ranked by \XPA, the dominant class equals \(y_t\) for 99.55\% and 98.18\% of samples respectively, confirming that \XPA closely tracks \CPA in the high-score regime where it matters most. We show additional results in \Cref{app:dps}.

To further confirm the practicality of \XPA, we randomly sample 200 target instances with $y_t = 0$ on CIFAR-10 and measure the Pearson correlation between ASR and each metric. Under GM, the correlations with ASR are $-0.144$ for \CPA and $-0.095$ for \XPA; under BP, they are $-0.511$ for \CPA and $-0.390$ for \XPA. Both metrics exhibit a consistent negative correlation with ASR, confirming that \XPA is a reliable label-agnostic surrogate for \CPA.

\subsection{Fine-grained Metrics: Poisoning Distance and Budget}
\label{sec:exp_delta}
\begin{table}[t]
    \centering
    \caption{Measuring the poisoning difficulty of GM on CIFAR-10 using $\delta$ and $\tau$ for the poison classes with the highest and lowest ASR over all target classes $y_t$. Anomaly cases where the prediction does not conform to ASR are marked with \underline{underline}.}
\scalebox{0.85}{\begin{tabular}{lccc|ccc}
\toprule
\multirow{2}{*}[-.6ex]{$y_t$} & \multicolumn{3}{c|}{\textbf{Lowest ASR $y_p$}} & \multicolumn{3}{c}{\textbf{Highest ASR $y_p$}} \\
\cmidrule(l{0pt}r{0pt}){2-7}
  & distance $\delta$ & budget $\tau$ & ASR & distance $\delta$ & budget $\tau$ & ASR \\
\midrule
plane & $0.119 \pm 0.042$ & $0.00237 \pm 0.00336$ & $0.57 \pm 0.42$ & $0.105 \pm 0.030$  & $0.00190 \pm 0.00321$ & $0.74 \pm 0.35$ \\
car & $0.124 \pm 0.040$ & $0.00171 \pm 0.00617$ & $0.83 \pm 0.28$ & $0.112 \pm 0.030$ & $0.00097 \pm 0.00187$ & $0.90 \pm 0.23$ \\
bird  & $0.114 \pm 0.035$ & $0.00237 \pm 0.00425$ & $0.60 \pm 0.42$ & $0.097 \pm 0.033$ & $0.00140 \pm 0.00229$ & $0.84 \pm 0.28$ \\
cat &  \underline{$0.093 \pm 0.031$} & $0.00049 \pm 0.00093$ & $0.92 \pm 0.21$ &  $0.095 \pm 0.026$  & $0.00043 \pm 0.00121$ & $0.99 \pm 0.05$ \\
deer & $0.123 \pm 0.042$ & $0.00221 \pm 0.00494$ & $0.86 \pm 0.26$ & $0.092 \pm 0.034$ & $0.00117 \pm 0.00180$ & $0.99 \pm 0.07$ \\
dog & $0.115 \pm 0.038$ & \underline{$0.00095 \pm 0.00128$} & $0.91 \pm 0.19$  & $0.102 \pm 0.033$ & $0.00246 \pm 0.00526$ & $0.98 \pm 0.07$ \\
frog & $0.137 \pm 0.042$ & $0.00197 \pm 0.00219$ & $0.83 \pm 0.26$ & $0.112 \pm 0.035$  & $0.00158 \pm 0.00240$ & $0.98 \pm 0.07$ \\
horse & \underline{$0.106 \pm 0.041$} & \underline{$0.00081 \pm 0.00227$} & $0.69 \pm 0.36$ & $0.137 \pm 0.048$ & $0.00123 \pm 0.00214$ & $0.87 \pm 0.24$ \\
ship  & $0.119 \pm 0.037$ & $0.00336 \pm 0.00497$ & $0.67 \pm 0.38$ & $0.085 \pm 0.031$ & $0.00232 \pm 0.00564$ & $0.93 \pm 0.19$ \\
truck & $0.125 \pm 0.033$ & $0.00130 \pm 0.00138$ & $0.80 \pm 0.28$ & $0.106 \pm 0.031$ & $0.00058 \pm 0.00097$ & $0.95 \pm 0.11$ \\
\bottomrule
\end{tabular}}
\label{tab:y_p_choice}
\end{table}

\begin{figure}[t!]
\centering
    \begin{subfigure}{0.32\textwidth}        \includegraphics[width=\linewidth]{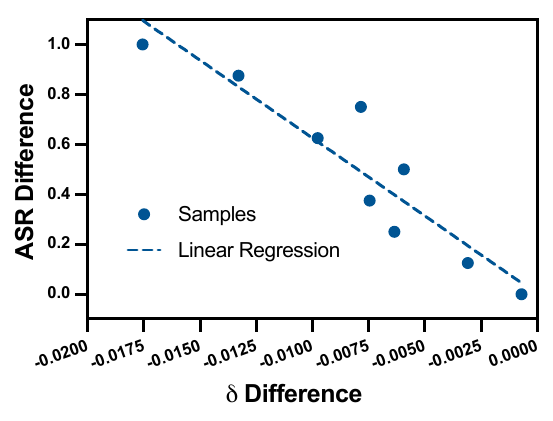}
        \caption{}
    \end{subfigure}
    \hfill
    \begin{subfigure}{0.32\textwidth}
        \includegraphics[width=1.1\linewidth]{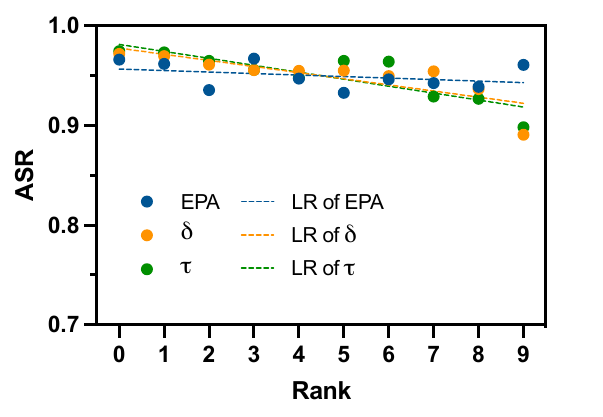}
        \caption{Low \CPA samples.}
    \end{subfigure}
    \hfill
    \begin{subfigure}{0.32\textwidth}
        \includegraphics[width=1.1\linewidth]{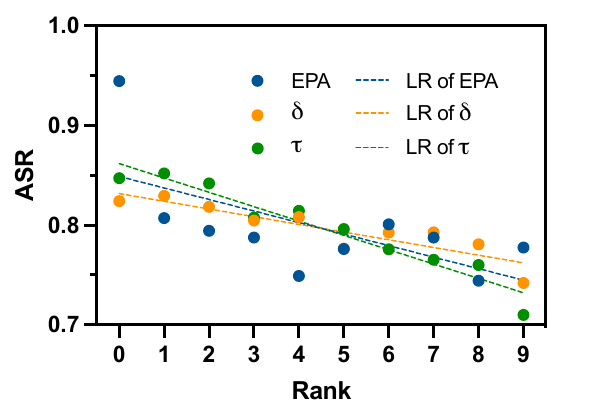}
        \caption{High \CPA samples.
        }
    \end{subfigure}

\caption{(a) Correlation between pairwise $\delta$ difference and ASR difference; (b) and (c) Comparison between all of our metrics for low/high \CPA samples. }
\label{fig:delta_tau_cpa}
\end{figure}

The previous methods are agnostic of the target poison label $y_p$, which appears to have a significant impact on poisoning difficulty (\Cref{fig:main_poison_class_differences_high_and_low_cpa}).
We now investigate the poisoning distance $\delta$ and the poison budget measure $\tau$, which take the poison label $y_p$ into account. We examine our prior results in the from-scratch setting.
First, we look at the choice of $y_p$ with respect to each target class. For each target class, we examine the same 100 test samples (50 highest/lowest \CPA targets) and calculate the average ASR for each $y_p$. We report the two $y_p$ classes with the lowest/highest average ASR and compare the average $\delta$ and $\tau$ values in \Cref{tab:y_p_choice}. We pre-screen out $(\xte, y_p)$ pairs (1135 out of 9000) where the clean model already predicts $y_p$, as these would trivially count as successful attacks without any poisoning. While we observe $\delta$ and $\tau$ are generally capable of identifying easy/hard poison classes, there are some anomalies: for example, the target classes dog and cat have a smaller difference in ASR between the highest/lowest ASR $y_p$, making them more difficult to differentiate.
To provide further understanding on the role of the poison class $y_p$, for every individual instance $\xte$, we enumerate the choice of $y_p$ and create pairs  $((\xte,y_p^1),(\xte,y_p^2))$ (there are 9 choose 2, which is 36 pairs in total). For each pair, we calculate its corresponding ASR difference and $\delta$ difference.
After obtaining all $1000 \times 36$ pairwise ASR and $\delta$ differences, we plot the correlation of the average $\delta$ difference with respect to the 9 possible ASR differences\footnote{Note that for each attack instance, as we perform 8 trials, ASR can take 9 values in $[0,1]$ with an interval of $1/8$. The ASR difference can only take the same 9 values as we restrict the difference to be positive.} in \Cref{fig:delta_tau_cpa}(a) and observe that $\delta$ is generally reliable even for differentiating pairwise differences.

Moreover, \CPA alone is not sufficient to further distinguish poisoning difficulty within groups of samples with similar \CPA values. In \Cref{fig:delta_tau_cpa}(b)(c), we rank samples in the high and low \CPA regions according to their \CPA, $\delta$, and $\tau$ into deciles and plot the average ASR for each decile. We observe that $\delta$ and $\tau$ cover a much wider range of ASR, demonstrating their ability to provide finer-grained predictions within groups that \CPA alone cannot resolve.

\paragraph{Practical guidelines.} Our metrics serve two concrete use cases. For evaluators, we recommend using \CPA to rank the test set and reporting ASR on the top-$k\%$ hardest-to-poison samples (highest \CPA) as a worst-case evaluation benchmark, rather than averaging over random samples. This gives a more conservative and informative picture of true attack effectiveness. For defenders, the same ranking identifies the most vulnerable samples (lowest \CPA), which should receive prioritized protection, for example through manual inspection, ensemble disagreement checks, or certified defenses. When the specific poison class matters (e.g., for high-stakes misclassification scenarios), $\delta$ and $\tau$ can further refine this prioritization at the poison-class level without requiring any actual attack to be run.

%% file: sections/ablation.tex
\subsection{Ablation studies}

\textbf{Necessity of training dynamics:}
Our metrics \CPA and \XPA operate on the principle that samples which the model confidently classifies are harder to poison.
A natural baseline is to use the model's final-epoch softmax confidence on the true label $y_t$ as a simpler proxy.
We run GM on CIFAR-10 with $y_t=\text{plane}, y_p=\text{bird}$ and report the ASR, \CPA, and confidence of $y_t$ for high and lower \CPA samples\footnote{We restrict the confidence of $y_t>0.98$ for samples with lower \CPA.} in \Cref{tab:base_confidence_ablation}.
We observe that this simpler confidence metric is not predictive of poisoning difficulty, in stark contrast to \CPA which captures predictions over the full course of training.

\begin{table}[h!]
\vspace{-.5em}
    \centering
    \caption{Ablation study on predicting poisoning difficulty with the confidence of $y_t$.}
    \scalebox{0.85}{\begin{tabular}{cccc}
    \toprule
    $\xte$ & confidence of $y_t$ & average \CPA & average ASR \\
    \midrule
    high \CPA & $0.9985 \pm 0.00309$ & $0.9955 \pm 0.00165$ & $0.468 \pm 0.3718$ \\
    lower \CPA & $0.9999 \pm 0.00003$ & $0.9249 \pm 0.03210$ & $0.905 \pm 0.1899$ \\
    \bottomrule
    \end{tabular}}
    \label{tab:base_confidence_ablation}
\end{table}

\textbf{Poison budget:} 
Recall that \CPA is agnostic of the poison budget $\epsilon$, and its predictive power may vary for different $\epsilon$.
As an extreme example, when $\epsilon=100\%$ (say), it should be easy to poison \emph{any} point, and the \CPA value is irrelevant.
In \Cref{tab:dog_budget_ablation_table}, we explore how \CPA predicts poisoning difficulty for successively reduced poisoning budgets.
Indeed, for relatively large poison budgets (e.g., $\epsilon=1\%$), poisoning attacks are easy and \CPA has limited discriminative value.
However, the predictive power of \CPA becomes more apparent for smaller budgets, where poisoning is non-trivial and the relative difficulty of poisoning different samples is a meaningful concept. We use the class ``dog'' as $y_t$ for this experiment, as it is generally easy to poison at $\epsilon=1\%$, making it a good stress test for \CPA at lower budgets. We provide results for a wider range of $\epsilon$ in Appendix \cref{app:epsilon_sweep}.

\begin{table}[h!]
\vspace{-.5em}
    \centering
      \caption{Ablation study on predicting poisoning difficulty with \CPA for various attack budget $\epsilon$.}
\scalebox{0.8}{\begin{tabular}{ cccccc } 
\toprule
$\xte$& $\epsilon=1\%$ & $\epsilon=0.75\%$ & $\epsilon=0.5\%$& $\epsilon=0.25\%$ & $\epsilon=0.1\%$ \\
\midrule
high \CPA  & $0.963 \pm 0.084$  & $0.963 \pm 0.119$  & $0.738 \pm 0.375$  & $0.588 \pm 0.382$ & $0.263 \pm 0.216$\\
low \CPA & $0.988 \pm 0.040$ & $1.000 \pm 0.000$ & $0.963 \pm 0.060$ & $0.950 \pm 0.121$ & $0.662 \pm 0.391$\\
\bottomrule
\end{tabular}}
\label{tab:dog_budget_ablation_table}
\end{table}

\textbf{Additional Datasets and Models}: 
We perform experiments on an alternative architecture, VGG-13 with CIFAR-10. In \Cref{tab:cifar10_epa_vgg13}, we confirm that \CPA continues to reliably predict poisoning difficulty on CIFAR-10.
\begin{table}[h!]
    \centering
    \vspace{-1em}
    \caption{ASR of poisoning attacks using \CPA on CIFAR-10 with VGG-13.}
\scalebox{0.8}{\begin{tabular}{ccc|cc}
\toprule
 \multirow{2}{*}[-.6ex]{$\xte$}
 & \multicolumn{2}{c|}{$y_p=\text{car}$} & \multicolumn{2}{c}{$y_p=\text{bird}$}\\
 \cmidrule(l{0pt}r{0pt}){2-5}
 & GM (ASR) & BP (ASR) & GM (ASR) & BP (ASR) \\
\midrule
high \CPA & 0.094 $\pm$ 0.104 & 0.000 $\pm$ 0.000 & 0.219 $\pm$ 0.256 & 0.000 $\pm$ 0.000 \\
middle \CPA & 0.969 $\pm$ 0.054 & 0.031 $\pm$ 0.054 & 0.781 $\pm$ 0.379 & 0.000 $\pm$ 0.000 \\
low \CPA & 1.000 $\pm$ 0.000 & 0.688 $\pm$ 0.400 & 1.000 $\pm$ 0.000 & 0.656 $\pm$ 0.389 \\
\bottomrule
\end{tabular}}
\label{tab:cifar10_epa_vgg13}
\end{table}

We further evaluate \CPA for predicting poisoning difficulty on TinyImageNet in \Cref{tab:tinyimagenet}. 
We select samples with the highest, middle, and lowest \CPA values from $y_t = 0$, and perform poisoning towards $y_p = 1, 2$ using GM and BP under a budget of $\epsilon = 0.05\%$.
We observe that \CPA does not consistently correlate with poisoning difficulty. We believe this failure is largely due to underfitting, as there are only 500 samples in each class. In such cases, prediction trajectories become noisy and poorly structured, making the training dynamics unreliable and diminishing the effectiveness of \CPA.

\begin{table}[h!]
    \centering
    \vspace{-1em}
    \caption{ASR of poisoning attacks using \CPA on TinyImageNet ($y_t=0$).}
\scalebox{0.8}{\begin{tabular}{ccc|cc}
\toprule
 \multirow{2}{*}[-.6ex]{$\xte$}
 & \multicolumn{2}{c|}{$y_p=1$} & \multicolumn{2}{c}{$y_p=2$}\\
 \cmidrule(l{0pt}r{0pt}){2-5}
 & GM (ASR) & BP (ASR) & GM (ASR) & BP (ASR) \\
\midrule
high \CPA & 0.656 $\pm$ 0.409 & 0.719 $\pm$ 0.223 & 0.781 $\pm$ 0.379 & 0.594 $\pm$ 0.368 \\
middle \CPA & 0.844 $\pm$ 0.205 & 0.625 $\pm$ 0.342 & 0.938 $\pm$ 0.062 & 0.688 $\pm$ 0.410 \\
low \CPA & 0.250 $\pm$ 0.354 & 0.344 $\pm$ 0.389 & 0.531 $\pm$ 0.285 & 0.750 $\pm$ 0.177 \\
\bottomrule
\end{tabular}}
\label{tab:tinyimagenet}
\end{table}

%% file: sections/conclusion.tex
\section{Conclusion}
\label{sec:conclusion}
In this paper, we argued that evaluating targeted data poisoning attacks by averaging success rates over random samples is misleading, as it conflates easy and hard targets and obscures true worst-case effectiveness. We introduced a two-level framework of metrics computable without executing any actual attacks. At the coarse level, \CPAfull~(\CPA) and \XPAfull~(\XPA) use clean training dynamics to separate vulnerable from robust samples. At the fine-grained level, poisoning distance $\delta$ and poison budget lower bound $\tau$ provide poison-class-specific predictions from clean model weights alone.
Our experiments confirm that attacks are substantially less effective on high-\CPA samples than average-case evaluations suggest, and that $\delta$ and $\tau$ further resolve difficulty within groups that \CPA cannot distinguish. These metrics translate directly into practice: evaluators should assess worst-case ASR on the hardest samples, and defenders should prioritize protection for the most vulnerable ones.
We discuss limitations and future directions in Appendix~\ref{app:limitations}.

\paragraph{Acknowledgements.} We gratefully acknowledge funding support from NSERC, the Canada CIFAR AI Chairs program, and an Ontario Early Researcher Award. Resources used in preparing this research were provided, in part, by the Province of Ontario, the Government of Canada through CIFAR, and companies sponsoring the Vector Institute.

%% file: sections/appendix.tex
\section{Limitations and Future Works}
\label{app:limitations}

\paragraph{Limitations.} As shown in our experiments, our metrics may occasionally yield inaccurate predictions of poisoning difficulty, indicating room for improvement. In particular, on more complex datasets such as TinyImageNet, both \CPA and \XPA become significantly less predictive. We hypothesize that this is due to increased class diversity and weaker feature separability, which lead to less stable prediction dynamics during clean training. This suggests that prediction-dynamics-based metrics may be less reliable in high-complexity regimes.

\paragraph{Future Works.} Several potential future directions emerge from our work: (1) Data-centric defenses that optimize test samples to defend against targeted data poisoning attacks — for example, defenders might apply carefully crafted adversarial noise to test data, similar to techniques used in adversarial examples; (2) Extending the framework to larger-scale and more complex settings, including foundation models and generative models (see Appendix~\ref{app:diffusion} for an initial exploration), and developing universal quantitative metrics for assessing poisoning difficulty across different model types.

\paragraph{Broader Impacts.} This work advances the evaluation standards for targeted data poisoning attacks by showing that average-case success rates over random samples obscure true worst-case vulnerability. Our metrics enable practitioners to identify the most vulnerable samples without running any attacks, providing a practical tool for proactive defense. We acknowledge a dual-use concern: the same metrics could help attackers select easy targets more efficiently. However, this risk is limited — identifying easy targets has always been possible by running multiple attacks; our metrics reduce the computational cost but do not introduce qualitatively new attack capabilities, and the defensive benefits are asymmetric relative to this marginal gain.

\section{Related Works}
\label{app:related_works}

\subsection{Data poisoning attacks}
\label{app:data-poisoning}
Data poisoning, an emerging training-time concern in modern ML pipelines, refers to the threat of (actively or passively) crafting ``poisoned'' training data $\Dpo$ so that systems trained on it (along with possibly clean in-house data $\Dcl$) are skewed toward certain behaviors. Significant research has been proposed to study the impact of such attacks on classification models.
For example, indiscriminate data poisoning \citep[\eg,][]{BiggioNL12,KL17,KohSL18,GonzalezBDPWLR17,DiakonikolasKKLSS19,LuKY22, LuKY23, LuWZY24, LuYKY24, LuYLKY24} is a general-purpose attack that aims to decrease the overall test accuracy. Similar formulations have been proposed for protecting user data \citep[\eg,][]{LiuC10,HuangMEBW21,YuZCYL21,FowlGCGCG21, FowlCGGBCG21,SadovalSGGGJ22,FuHLST21}. While data poisoning attacks can also involve testing-time manipulation—such as backdoor attacks \citep[\eg,][]{GuDG17,TranLM18,ChenLLLS17,SahaSP20} that aim to trigger malicious model behavior with particular patterns on test samples, we focus exclusively on training-time attacks in this paper.

Targeted attacks are insidious as they do not cause significant performance degradation (hence harder to detect) while still capable of causing system failure on targeted test instances. Previous works \citep{ShafahiHNSSDG18, GeipingFHCTMG20,ZhuHLTSG19e} have demonstrated the efficacy of such attacks against deep neural networks, reporting high attack success rates. However, these reported success rates are typically calculated by averaging over randomly selected test samples \citep{SchwarzschildGGDG20,GeipingFHCTMG20}. This aggregated metric fails to capture the instance-level difficulty of targeted data poisoning attacks, a critical gap we aim to address in this work. 

\subsection{Adding attack vs Replacing attack}
\label{app:add_replacing}
Realistically, an attacker would have no control on the clean set $\Dcl$, and data poisoning attacks \citep[\eg,][]{BiggioNL12,KohSL18,GonzalezBDPWLR17,LuKY22, LuKY23} usually consider \emph{adding-only} attack where $\Dcl$ is intact and the size of $\Dtr$ increases. However, targeted attacks \citep[\eg,][]{ShafahiHNSSDG18,AghakhaniMWKV20,GuoL20,ZhuHLTSG19e} consider \emph{replacing} attacks where part of the clean set $\Dcl$ is substituted with $\Dpo$ while the size of $\Dtr$ is unchanged. Note that such a substitution is performed by simply adding noise to the original clean samples. Such a setting could resonate in targeted settings as it would keep the balance between classes. In this paper, we follow previous works and consider \emph{replacing attacks}. While the practicality of such attacks are beyond our scope, we note that the key technical differences comparing with adding-only attacks: replacing attacks are notably easier as it reduces $|\Dcl|$ and considers a slightly higher $\epsilon$ as $|\Dtr|$ is a constant (see Appendix C.10 in \citep{LuKY23} for a detailed discussion).

\subsection{Measuring classification difficulty}
\label{app:class_diff}
The problem of measuring classification difficulty has been explored in prior literature. For instance, \citet{AgarwalDH22} proposed variance of gradient (VOG) as a method to rank examples by classification difficulty. VOG could potentially serve as an alternative to \CPA for measuring classification difficulty in \Cref{assump:classification} and may function as an indicator for predicting poisoning vulnerability—a direction we intend to investigate in future work. Additionally, out-of-distribution (OoD) detection techniques such as PCA \citep{hawkins1974detection} and KDE \citep{davis2011remarks} could potentially identify \emph{hard-to-classify} (and possibly \emph{easy-to-poison}) anomalous samples.

Furthermore, our approach relates to selective classification~\cite{RabanserTHDP22}, where models reject inputs likely to be misclassified while maintaining high performance on accepted inputs. Specifically, \citet{RabanserTHDP22} leverages prediction agreement between intermediate training stages and the final epoch—a strategy similar to our \CPA metric that also analyzes clean training dynamics. However, unlike selective classification, we assign a \CPA
 score to every test instance rather than implementing a rejection mechanism.

\subsection{Connection and differences with model-targeted attacks}
\label{app:MTA}
We note that our core idea of poisoning distance is closely related to model-targeted indiscriminate attacks which we denote as MTA \citep{SuyaMSET21,LuKY23,KohSL18}, where these attacks consider a set of target parameters $\wv_p$ as the target and apply gradient-based poisoning attacks to achieve $\wv_p$. While the concept of target parameters is also used in our paper, we emphasize key differences: (1) \emph{Task}: MTA considers $\wv_p$ to be a model with low test accuracy, which can be generated with a gradient-based parameter corruption attack \citep{SunZRLL20}. We consider a set of $\wv_p$ that only misclassifies one single test sample. (2)  \emph{Using $\wv_p$}: MTA uses $\wv_p$ as the endgoal to generate poisoning attacks, we use $\wv_p$ as proxies to quantify poisoning difficulty. (3) \emph{Attack vs Defense}: MTA are designed for more effective attacks, while our algorithm estimates $\wv_p$ and $\delta$ to help practitioners understand targeted attack difficulties and design better defenses.

\section{Algorithm on Estimating Poisoning Distance}
\label{app:algorithms}

Recall that in \Cref{sec:delta} we propose to use a binary search based algorithm to estimate the poisoning distance $\delta$. We present the algorithm in \Cref{alg:delta}.

\begin{algorithm}[ht]
\DontPrintSemicolon
    \KwIn{clean parameters $\wv_c$, target $\xte$, poison label $y_p$, precision parameter $\alpha = 10^{-4}$}
    calculate the gradient $\gv = \nabla_{\wv_c} \ell(\xte; \wv_c, y_p)$

    instantiate the upper bound $u = \infty$, lower bound $l = 0$, and medium $m = 0.5$ for binary search

    \While{$u - l > \alpha$} {
        \If{$u = \infty$} {
            set $m = 2m$
        }
        \Else {
            set $m = \frac{u + l}{2}$
        }

        \If{$f(\xte; \wv_c - m \cdot \gv) = y_p$} {
            set $u = m$
        }
        \Else {
            set $l = m$
        }
    }

\textbf{return} the estimated poisoning distance $\delta=u$

\caption{Poisoning Distance Estimation}
\label{alg:delta}
\end{algorithm}
\vspace{-1em}

\section{Additional Experiments on Classification Models}

\subsection{Computing resource \& time}

\label{app:compute}
\textbf{Targeted attacks:} Due to the extensive number of attacks conducted, we distributed our experiments across three distinct clusters equipped with NVIDIA 4090 (cluster 1), A100 (cluster 2), and RTX6000 GPUs (cluster 3). The computational requirements varied significantly by task: training models from scratch (GM experiments) required up to 1 hour 40 minutes on all clusters for CIFAR-10/ResNet-18 configurations, while TinyImageNet/VGG16 experiments (GM and BP) demanded up to 3 hours 10 minutes on cluster 2. Transfer learning experiments were considerably more efficient, requiring only 66-72 seconds for BP and 60-63 seconds for FC on clusters 2 and 3, respectively.

\textbf{Measuring poisoning difficulty:} We conducted all experiments on the NVIDIA 4090 cluster. For \CPA calculations, the computational cost scales linearly with the number of trials $M$ multiplied by the clean training time. For individual test samples, computing all nine possible $\delta$ values for a single set of $\wv_c$ requires just 1.3 seconds, while calculating all nine possible $\tau$ values takes approximately 30 seconds on our ResNet-18/CIFAR-10 experimental setup.

\subsection{Additional experiments on transfer learning}
\label{add_exp:transfer}
For the transfer learning setting, we start from a pre-trained model with reasonable performance and visualize the model change after an attack.
We report the average change of confidence for $y_p$ and $y_t$ for each test sample and confirm that \CPA is a reliable metric to measure poisoning difficulty.
Moreover, to check whether \CPA is a reliable tool for predicting attack success, we report the average \CPA of test targets that are successfully and unsuccessfully poisoned in \Cref{tab:success_vs_failure_transfer_learning}. We observe that \CPA is capable of clearly differentiating between successful attacks and failed attacks in most cases, while the prediction region may occasionally overlap.  

\begin{table}[t]
    \caption{The ASR, change of confidence for $y_p$ and $y_t$ before/after attack for high/low \CPA test samples with FC and BP attack on CIFAR-10 with transfer learning.}
    \centering
        \scalebox{0.9}{\begin{tabular}{ccc|cc|cc}
        \toprule
        \multirow{2}{*}[-.6ex]{\textbf{$\xte$}}& \multicolumn{2}{c|}{\textbf{ASR}} & \multicolumn{2}{c|}{\textbf{change of confidence ($y_p$)}} & \multicolumn{2}{c}{\textbf{change of confidence ($y_t$)}} \\
        \cmidrule(l{0pt}r{0pt}){2-7}
         & FC & BP & FC & BP & FC & BP \\
        \midrule
        high \CPA & 0.012 & 0.498 & $0.040 \pm 0.031$ & $0.479 \pm 0.133$ & $-0.048 \pm 0.036$ & $-0.503 \pm 0.137$ \\
        low \CPA & 0.284 & 0.947 & $0.156 \pm 0.041$ & $0.661 \pm 0.063$ & $-0.198 \pm 0.053$ & $-0.750 \pm 0.058$ \\
        \bottomrule
        \end{tabular}}
    \label{tab:easy_vs_hard_transfer_learning}
    \vspace{-1em}
\end{table}

\begin{table}[h!]
    \centering
        \caption{The average \CPA and confidence of $y_p$ after clean training for successful/failed attacks using the FC and BP attack on CIFAR-10 with transfer learning.}
    \scalebox{0.85}{\begin{tabular}{ccc|cc}
    \toprule
     \multirow{2}{*}[-.6ex]{\textbf{Attack Success}}& \multicolumn{2}{c|}{\textbf{Average \CPA}} & \multicolumn{2}{c}{\textbf{Confidence of $y_p$}} \\
    \cmidrule(l{0pt}r{0pt}){2-5}
     & FC & BP & FC & BP \\
    \midrule
    \cmark & $0.411 \pm 0.116$ & $0.589 \pm 0.255$ & $0.142 \pm 0.092$ & $0.043 \pm 0.074$ \\
    \xmark & $0.725 \pm 0.264$ & $0.912 \pm 0.148$ & $0.012 \pm 0.033$ & $0.001 \pm 0.002$ \\
    \bottomrule
    \end{tabular} }   \label{tab:success_vs_failure_transfer_learning}
\end{table}

\subsection{Additional experiments on \XPA}
\label{app:dps}
Recall that \XPA measures the fraction of training-time prediction assigned to the \emph{dominant label} instead of the ground-truth label \(y_t\). To evaluate whether \XPA can serve as a practical surrogate for \CPA, we compute \XPA for all CIFAR-10 test samples under the same clean-training setup and rank the samples by their \XPA scores. For each prefix of this ranking, we measure the proportion of samples whose dominant class coincides with the ground-truth label \(y_t\). We observe that \XPA is highly consistent with \CPA: among the top 20\% and top 40\% of samples ranked by \XPA, the dominant class equals \(y_t\) for 99.55\% and 98.18\% of samples, respectively, demonstrating \XPA is a strong surrogate of \CPA.

\begin{table}[h!]
    \centering
    \caption{Agreement between \XPA and \CPA on CIFAR-10. For each prefix of the test samples ranked by \XPA, we report the proportion of samples whose dominant class equals the ground-truth label \(y_t\).}
    \label{tab:xpa_surrogate}
    \scalebox{0.95}{
    \begin{tabular}{lccc}
        \toprule
        \textbf{Prefix} & \textbf{\# Samples} & \textbf{\# Dominant = \(y_t\)} & \textbf{Proportion} \\
        \midrule
        Top 20\%  & 2000  & 1991 & 0.9955 \\
        Top 40\%  & 4000  & 3927 & 0.9818 \\
        Top 60\%  & 6000  & 5705 & 0.9508 \\
        Top 80\%  & 8000  & 7113 & 0.8891 \\
        Top 100\% & 10000 & 8024 & 0.8024 \\
        \bottomrule
    \end{tabular}}
\end{table}

\subsection{Effect of poison budget on targets with different \CPA}
\label{app:epsilon_sweep}

We investigate how the poisoning budget $\epsilon$ interacts with target-specific difficulty. 
Specifically, we select the top-$n$ high-\CPA (hard) and bottom-$n$ low-\CPA (easy) samples on CIFAR-10 ($n=10$), evaluate GM attacks under varying budgets, and record the ASR in each case.
As shown in Table~\ref{tab:epsilon_sweep}, we sweep $\epsilon$ around a reference point $\epsilon=0.01$: for high-\CPA targets we increase $\epsilon$, while for low-\CPA targets we decrease $\epsilon$.
We observe a clear asymmetry. Increasing $\epsilon$ improves attack success on high-\CPA targets, but does not fully close the gap with low-\CPA targets. In contrast, low-\CPA targets remain highly vulnerable even as $\epsilon$ decreases.
This suggests that poisoning difficulty is largely intrinsic to the target, and cannot be fully compensated by increasing the attack budget.

\begin{table}[h!]
\centering
\caption{Effect of poisoning budget $\epsilon$ on targets with different \CPA on CIFAR-10.}
\label{tab:epsilon_sweep}
\scalebox{0.85}{
\begin{tabular}{c|ccccccccc}
\toprule
\diagbox{$\xte$}{$\epsilon$}
& 0.1\% & 0.25\% & 0.5\% & 0.75\% & 1\% & 1.5\% & 2\% & 3\% & 5\% \\
\midrule
high \CPA
& / & / & / & / & 0.638 & 0.700 & 0.775 & 0.713 & 0.725 \\
low \CPA
& 0.700 & 0.975 & 0.975 & 0.950 & 1.000 & / & / & / & / \\
\bottomrule
\end{tabular}}
\end{table}

\section{Targeted Attacks on Latent Diffusion Models}
\label{app:diffusion}

A key finding of this paper is that poisoning difficulty varies substantially across target samples, and can be predicted from target-specific properties without running any attacks. In this appendix, we ask whether the same holds for targeted attacks on latent diffusion models \cite{RombachBLEO22}. We consider the disguised copyright infringement (DCI) attack \cite{LuYLKY24}, where an attacker mimics the style of a copyrighted image $x_c$ via a disguise image $x_d$ crafted from a base image $x_b$. We find that attack success depends strongly on two target-specific structural factors — the appearance of $x_b$ and the complexity of $x_c$ — suggesting that instance-level difficulty is a general phenomenon beyond classification models.

\begin{figure*}[hbt!]
    \centering
    \setlength{\tabcolsep}{2pt}
    \small
    \begin{tabular}{cccc}
        \textbf{Base $x_b$} & \textbf{Disguise $x_d$} & \textbf{$x_c$} & \textbf{Generated by textual inversion} \\[4pt]
        \includegraphics[width=0.14\textwidth]{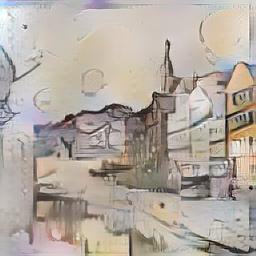} &
        \includegraphics[width=0.14\textwidth]{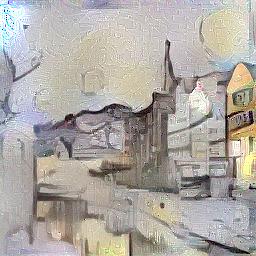} &
        \includegraphics[width=0.14\textwidth]{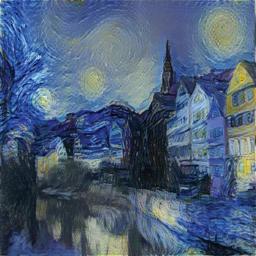} &
        \includegraphics[width=0.54\textwidth]{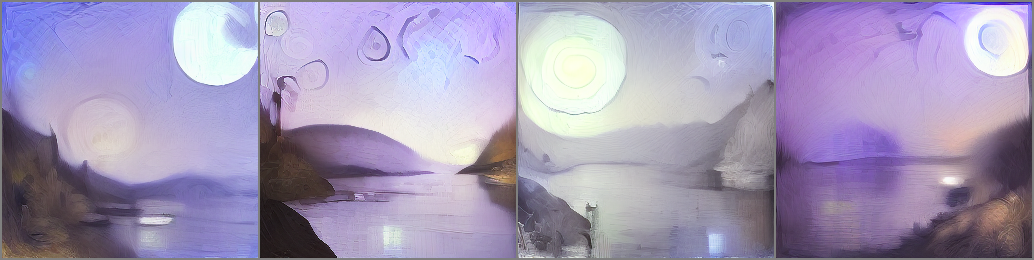} \\
        \multicolumn{4}{l}{(a) Original $x_b$} \\[4pt]
        \includegraphics[width=0.14\textwidth]{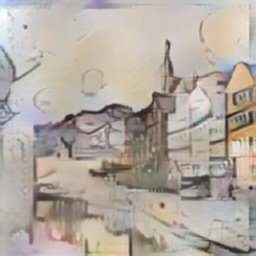} &
        \includegraphics[width=0.14\textwidth]{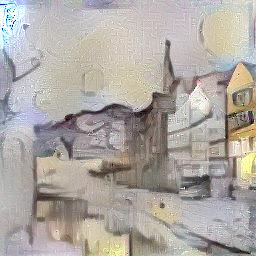} &
        \includegraphics[width=0.14\textwidth]{img/appendix_b/blur/van_gogh_rs.jpg} &
        \includegraphics[width=0.54\textwidth]{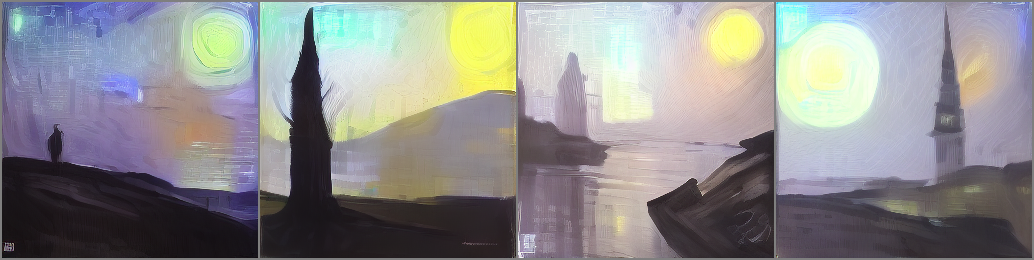} \\
        \multicolumn{4}{l}{(b) Blurry $x_b$, kernel size = 3} \\[4pt]
        \includegraphics[width=0.14\textwidth]{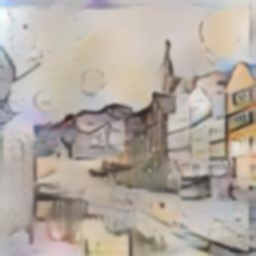} &
        \includegraphics[width=0.14\textwidth]{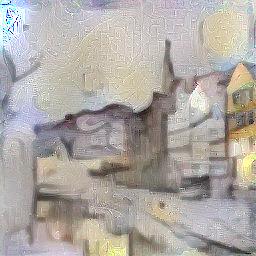} &
        \includegraphics[width=0.14\textwidth]{img/appendix_b/blur/van_gogh_rs.jpg} &
        \includegraphics[width=0.54\textwidth]{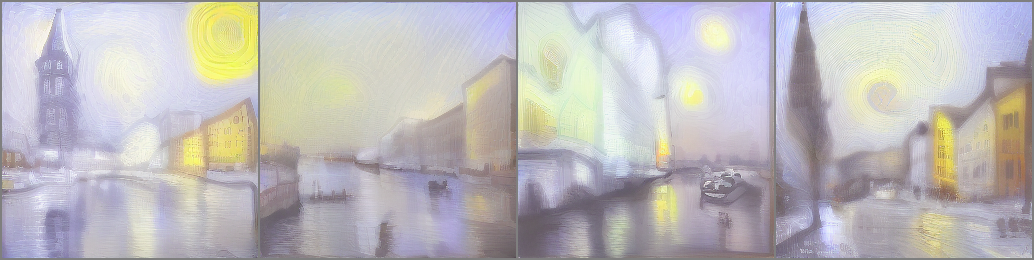} \\
        \multicolumn{4}{l}{(c) Blurry $x_b$, kernel size = 7} \\[4pt]
        \includegraphics[width=0.14\textwidth]{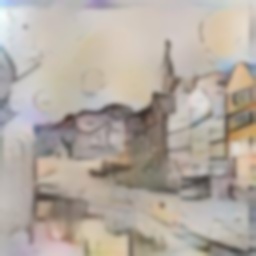} &
        \includegraphics[width=0.14\textwidth]{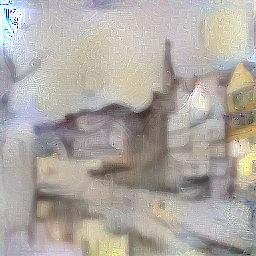} &
        \includegraphics[width=0.14\textwidth]{img/appendix_b/blur/van_gogh_rs.jpg} &
        \includegraphics[width=0.54\textwidth]{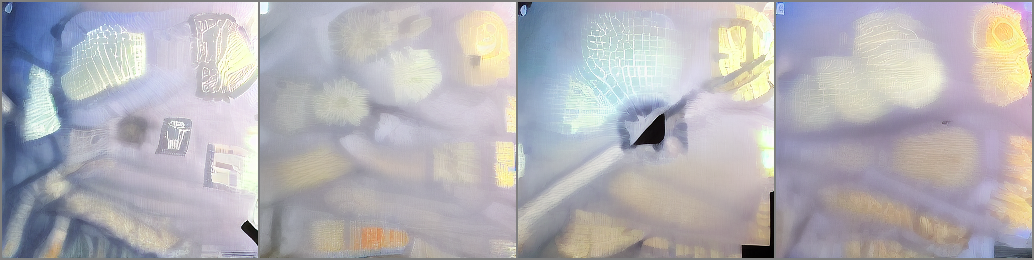} \\
        \multicolumn{4}{l}{(d) Blurry $x_b$, kernel size = 13} \\[4pt]
        \includegraphics[width=0.14\textwidth]{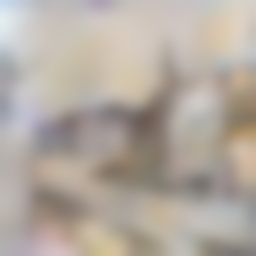} &
        \includegraphics[width=0.14\textwidth]{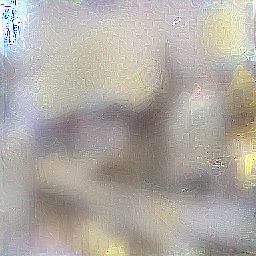} &
        \includegraphics[width=0.14\textwidth]{img/appendix_b/blur/van_gogh_rs.jpg} &
        \includegraphics[width=0.54\textwidth]{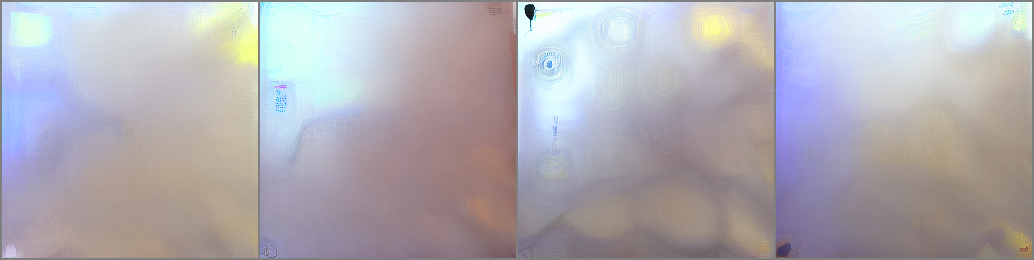} \\
        \multicolumn{4}{l}{(e) Blurry $x_b$, kernel size = 49} \\
    \end{tabular}
    \caption{Effect of Gaussian blur on $x_b$ for disguised copyright infringement. Each row shows the base image $x_b$, disguise $x_d$, copyright image $x_c$, and images generated by textual inversion after training on $x_d$. As kernel size increases, the generated images lose the characteristic style of $x_c$.}
    \label{fig:xb_blur_combined}
\end{figure*}

\textbf{Structure of $x_b$:} We follow the implementation of \cite{LuYLKY24}\footnote{\url{https://github.com/watml/disguised_copyright_infringement}} and consider the task of disguising style. We pick the drawing: The Neckarfront in Tubingen, Germany (photo by Andreas Praefcke) in the style of \emph{The Starry Night}, generated with Neural Style Transfer \citep{GatysEB15} as $x_c$. The base image $x_b$ is $x_c$ with another style (watercolor), generated with AdaIN-based \citep{HuangB17} style transfer.\footnote{\url{https://github.com/tyui592/AdaIN_Pytorch}} The disguise $x_d$ is generated using Algorithm 1 in \cite{LuYLKY24} and we train the disguise $x_d$ using textual inversion \cite{GalAAPBCC22} for generation.
We fix $x_c$ and study the role of the structure of $x_b$ by applying Gaussian blur with different kernel sizes (a larger kernel size results in a more blurry image). We report our results in \Cref{fig:xb_blur_combined}. We observe that by increasing the kernel size, the cirrus effect of the generated images dramatically decreases. When the kernel size is bigger than 10, the textual inversion model cannot learn any useful information. We conclude that preserving the structure of $x_b$ is essential for a successful data poisoning attack, highlighting the role of the poison image's appearance in poisoning difficulty.

\begin{figure*}[ht]
\vspace{-10pt}
    \centering
    \subfloat[{Base $x_b$}]{{\includegraphics[width=0.2\textwidth]{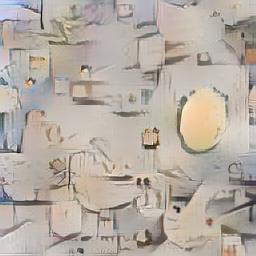}}}
    \subfloat[{Disguise $x_d$}]{{\includegraphics[width=0.2\textwidth]{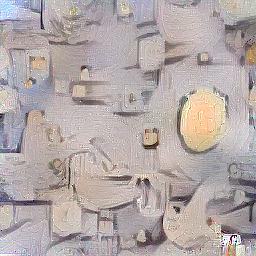}}}
    \subfloat[{$x_c$}]{{\includegraphics[width=0.2\textwidth]{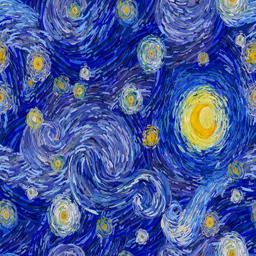}}}\\
    \subfloat[Images generated by textual inversion after training on the $x_d$]{\includegraphics[width=0.8\textwidth]{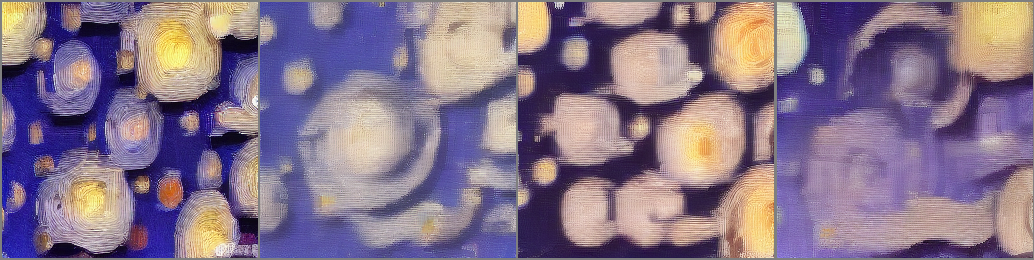}}
    \caption{Disguised copyright infringement with a different choice of $x_c$ (simpler layout). Style mimicry is unsuccessful, validating that the structure of $x_c$ also affects poisoning difficulty.}
    \label{fig:xc_structure}
\end{figure*}

\textbf{Structure of $x_c$:} We also observe that the structure of $x_c$ (the copyright image) affects poisoning difficulty. In \Cref{fig:xc_structure}, we choose another $x_c$\footnote{https://stock.adobe.com/images/glowing-moon-on-a-blue-sky-abstract-background-seamless-vector-pattern-in-the-style-of-impressionist-paintings/475101004} with a much simpler layout in the same style as \emph{The Starry Night}. We observe that style mimicry is unsuccessful for this poison instance, validating that poisoning success is also correlated with the structure of $x_c$.

